\documentclass[10pt,twocolumn,letterpaper]{article}

\usepackage{cvpr}
\usepackage{times}
\usepackage{epsfig}
\usepackage{graphicx}
\usepackage{amsmath}
\usepackage{amssymb}
\usepackage{comment}
\usepackage{url}
\usepackage{color}

\usepackage{pifont}
\usepackage{multirow}
\usepackage{makecell}
\usepackage{colortbl}
\usepackage{setspace}

\usepackage[
  pagebackref=true,
  breaklinks=true,
  colorlinks,
  bookmarks=false,
  citecolor=red,
  linkcolor=blue]{hyperref}

\newcommand{\cmark}{\ding{51}}%

\newcommand{\thickhline}{%
    \noalign {\ifnum 0=`}\fi \hrule height 1pt
    \futurelet \reserved@a \@xhline
}
\newcolumntype{x}[1]{>{\small\centering\let\newline\\\arraybackslash\hspace{0pt}}m{#1}}
\newcolumntype{L}[1]{>{\raggedright\let\newline\\\arraybackslash\hspace{0pt}}m{#1}}
\newcolumntype{C}[1]{>{\centering\let\newline\\\arraybackslash\hspace{0pt}}m{#1}}
\newcolumntype{R}[1]{>{\raggedleft\let\newline\\\arraybackslash\hspace{0pt}}m{#1}}

\makeatletter
\newcommand{\textsubsuperscript}[2]{%
  \begingroup
    \settowidth{\@tempdima}{\textsubscript{#1}}%
    \settowidth{\@tempdimb}{\textsuperscript{#2}}%
    \ifdim\@tempdima<\@tempdimb
      \setlength{\@tempdima}{\@tempdimb}%
    \fi
    \makebox[\@tempdima][l]{%
      \rlap{\textsubscript{#1}}\textsuperscript{#2}}%
  \endgroup}
\makeatother

\cvprfinalcopy 

\def\cvprPaperID{458} 

\ifcvprfinal\fi
\begin{document}

\title{Iterative Visual Reasoning Beyond Convolutions}

\author{Xinlei Chen\textsuperscript{1} \qquad Li-Jia Li\textsuperscript{2} \qquad Li Fei-Fei\textsuperscript{2} \qquad Abhinav Gupta\textsuperscript{1} \\
\textsuperscript{1}Carnegie Mellon University \qquad \textsuperscript{2}Google
}

\maketitle

\begin{abstract}
   We present a novel framework for iterative visual reasoning. Our framework goes beyond current recognition systems that lack the capability to reason beyond stack of convolutions. The framework consists of two core modules: a local module that uses spatial memory~\cite{chen2017spatial} to store previous beliefs with parallel updates; and a global graph-reasoning module. Our graph module has three components: a) a knowledge graph where we represent classes as nodes and build edges to encode different types of semantic relationships between them; b) a region graph of the current image where regions in the image are nodes and spatial relationships between these regions are edges; c) an assignment graph that assigns regions to classes. Both the local module and the global module roll-out iteratively and cross-feed predictions to each other to refine estimates. The final predictions are made by combining the best of both modules with an attention mechanism. We show strong performance over plain ConvNets, \eg achieving an $8.4\%$ absolute improvement on ADE~\cite{zhou2016semantic} measured by per-class average precision. Analysis also shows that the framework is resilient to missing regions for reasoning. 
\end{abstract}

\vspace{-0.05in}
\section{Introduction}
\vspace{-0.05in}
In recent years, we have made significant advances in standard recognition tasks such as image classification~\cite{he2016deep}, detection~\cite{ren2015faster} or segmentation~\cite{chen2016attention}. Most of these gains are a result of using feed-forward end-to-end learned ConvNet models. Unlike humans where visual reasoning about the space and semantics is crucial~\cite{biederman1982scene}, our current visual systems lack any context reasoning beyond convolutions with large receptive fields. Therefore, a critical question is how do we incorporate both \emph{spatial} and \emph{semantic} reasoning as we build next-generation vision systems.

\begin{figure}[t]
\centering
\includegraphics[width=0.9\linewidth]{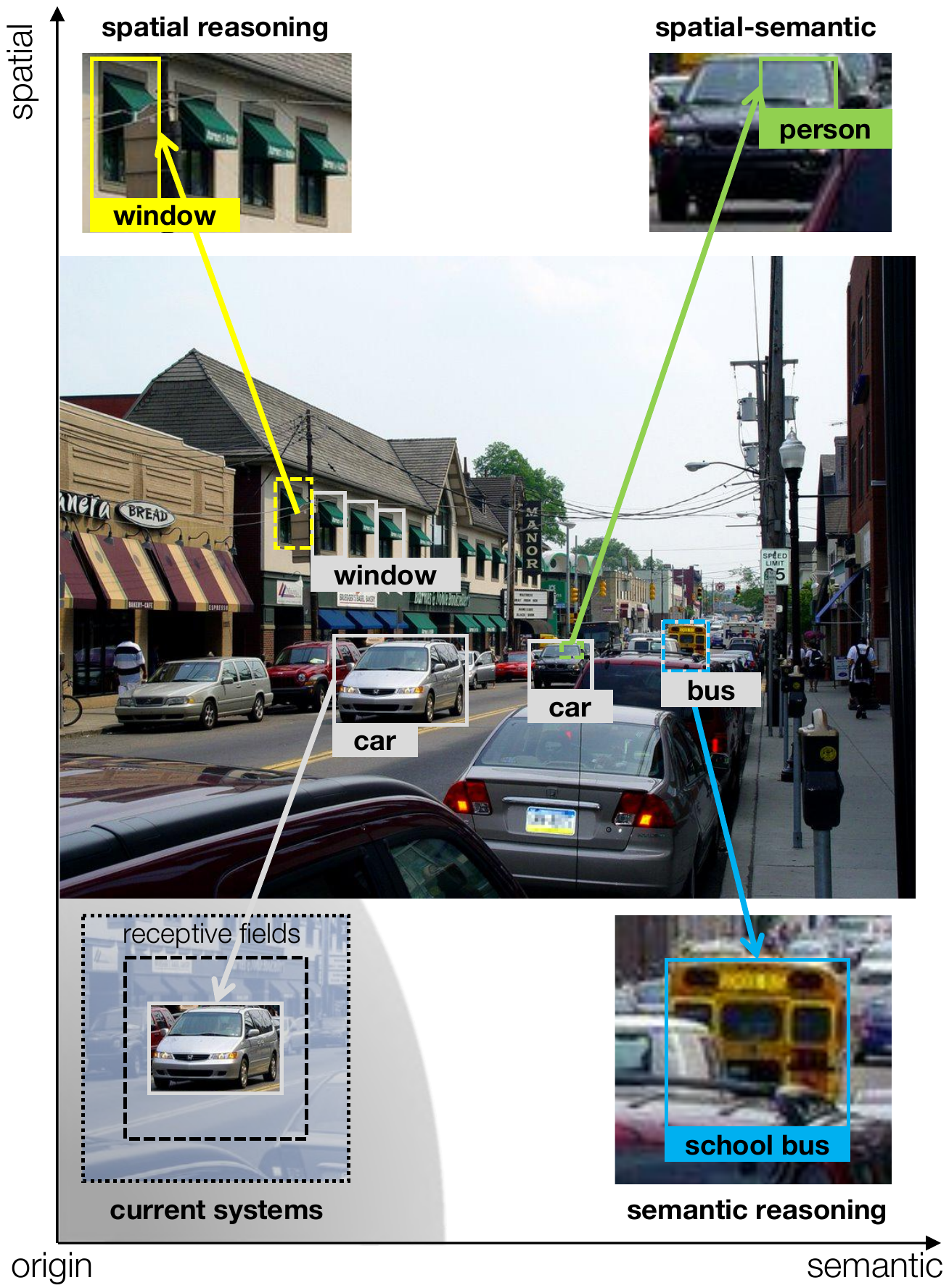}
\caption{{\small Current recognition systems lack the reasoning power beyond convolutions with large receptive fields, whereas humans can explore the rich space of spatial and semantic relationships for reasoning: \eg inferring the fourth ``window'' even with occlusion, or the ``person'' who drives the ``car''. To close this gap, we present a generic framework that also uses relationships to iteratively reason and build up estimates.}\label{fig:teaser}}
\vspace{-0.2in}
\end{figure}

Our goal is to build a system that can not only extract and utilize hierarchy of convolutional features, but also improve its estimates via spatial and semantic relationships. But what are spatial and semantic relationships and how can they be used to improve recognition? Take a look at Fig.~\ref{fig:teaser}. An example of spatial reasoning (top-left) would be: if three regions out of four in a line are ``window'', then the fourth is also likely to be ``window''. An example of semantic reasoning (bottom-right) would be to recognize ``school bus'' even if we have seen few or no examples of it -- just given examples of ``bus'' and knowing their connections. Finally, an example of spatial-semantic reasoning could be: recognition of a ``car'' on road should help in recognizing the ``person'' inside ``driving'' the ``car''.

A key recipe to reasoning with relationships is to \emph{iteratively} build up estimates. 
Recently, there have been efforts to incorporate such reasoning via top-down modules~\cite{ronneberger2015u,wei2016convolutional} or using explicit memories~\cite{xiong2016dynamic,marino2016more}. In the case of top-down modules, high-level features which have class-based information can be used in conjunction with low-level features to improve recognition performance. An alternative architecture is to use explicit memory. For example, Chen \& Gupta~\cite{chen2017spatial} performs sequential object detection, where a \emph{spatial memory} is used to store previously detected objects, leveraging the power of ConvNets for extracting dense context patterns beneficial for follow-up detections. 

However, there are two problems with these approaches: a) both approaches use stack of convolutions to perform \emph{local} pixel-level reasoning~\cite{divvala2009empirical}, which can lack a \emph{global} reasoning power that also allows regions farther away to directly communicate information; b) more importantly, both approaches assume enough examples of relationships in the training data -- so that the model can learn them from scratch, but as the relationships grow exponentially with increasing number of classes, there is not always enough data. A lot of semantic reasoning requires learning from few or no examples~\cite{fei2006one}. Therefore, we need ways to exploit additional structured information for visual reasoning.

In this paper, we put forward a generic framework for both spatial and semantic reasoning. Different from current approaches that are just relying on convolutions, our framework can also learn from structured information in the form of knowledge bases~\cite{chen2013neil,zhu2015building} for visual recognition. The core of our algorithm consists of two modules: the local module, based on spatial memory~\cite{chen2017spatial}, performs pixel-level reasoning using ConvNets. We make major improvements on efficiency by parallel memory updates. Additionally, we introduce a global module for reasoning beyond local regions. In the global module, reasoning is based on a \emph{graph} structure. It has three components: a) a knowledge graph where we represent classes as nodes and build edges to encode different types of semantic relationships; b) a region graph of the current image where regions in the image are nodes and spatial relationships between these regions are edges; c) an assignment graph that assigns regions to classes. Taking advantage of such a structure, we develop a reasoning module specifically designed to pass information on this graph. Both the local module and the global module roll-out iteratively and cross-feed predictions to each other in order to refine estimates. Note that, local and global reasoning are not isolated: a good image understanding is usually a compromise between background knowledge learned \emph{a priori} and image-specific observations. Therefore, our full pipeline joins force of the two modules by an attention~\cite{chen2016attention} mechanism allowing the model to rely on the most relevant features when making the final predictions.

We show strong performance over plain ConvNets using our framework. For example, we can achieve $8.4\%$ absolute improvements on ADE~\cite{zhou2016semantic} measured by per-class average precision, where by simply making the network deeper can only help ${\sim}1\%$. 

\vspace{-0.05in}
\section{Related Work}
\vspace{-0.05in}
\noindent{\bf Visual Knowledge Base.} Whereas past five years in computer vision will probably be remembered as the successful resurgence of neural networks, acquiring visual knowledge at a large scale -- the simplest form being labeled instances of objects~\cite{russakovsky2015imagenet,lin2014microsoft}, scenes~\cite{zhou2016semantic}, relationships~\cite{krishna2016visual} \etc -- deserves at least half the credit, since ConvNets hinge on large datasets~\cite{chensun2017}. Apart from providing labels using crowd-sourcing, attempts have also been made to accumulate structured knowledge (\eg relationships~\cite{chen2013neil}, $n$-grams~\cite{divvala2014learning}) automatically from the web. However, these works fixate on building knowledge bases rather than using knowledge for reasoning. Our framework, while being more general, is along the line of research that applies visual knowledge base to end tasks, such as affordances~\cite{zhu2015building}, image classification~\cite{marino2016more}, or question answering~\cite{wu2016ask}.

\noindent{\bf Context Modeling.} Modeling context, or the interplay between scenes, objects and parts is one of the central problems in computer vision. While various previous work (\eg scene-level reasoning~\cite{torralba2003context}, attributes~\cite{farhadi2009describing,parikh2011relative}, structured prediction~\cite{krahenbuhl2011efficient,desai2011discriminative,tu2010auto}, relationship graph~\cite{johnson2015image,lu2016visual,xu2017scene}) has approached this problem from different angles, the breakthrough comes from the idea of feature learning with ConvNets~\cite{he2016deep}. On the surface, such models hardly use any explicit context module for reasoning, but it is generally accepted that ConvNets are extremely effective in aggregating local pixel-to-level context through its ever-growing receptive fields~\cite{zeiler2014visualizing}. Even the most recent developments such as top-down module~\cite{xie2016top,lin2016feature,tdm_cvpr17}, pairwise module~\cite{santoro2017simple}, iterative feedback~\cite{wei2016convolutional,newell2016stacked,carreira2016human}, attention~\cite{yang2016stacked}, and memory~\cite{xiong2016dynamic,chen2017spatial} are motivated to leverage such power and depend on variants of convolutions for reasoning. Our work takes an important next step beyond those approaches in that it also incorporates learning from structured visual knowledge bases directly to reason with spatial and semantic relationships.

\noindent{\bf Relational Reasoning.} The earliest form of reasoning in artificial intelligence dates back to symbolic approaches~\cite{newell1980physical}, where relations between abstract symbols are defined by the language of mathematics and logic, and reasoning takes place by deduction, abduction~\cite{hobbs1988interpretation}, \etc. However, symbols need to be grounded~\cite{harnad1990symbol} before such systems are practically useful. Modern approaches, such as path ranking algorithm~\cite{lao2011random}, rely on statistical learning to extract useful patterns to perform relational reasoning on structured knowledge bases. As an active research area, there are recent works also applying neural networks to the graph structured data~\cite{scarselli2009graph,henaff2015deep,li2015gated,kipf2016semi,niepert2016learning,das2016chains,marino2016more}, or attempting to regularize the output of networks with relationships~\cite{deng2014large} and knowledge bases~\cite{hu2016deep}. However, we believe for visual data, reasoning should be both local and global: discarding the two-dimensional image structure is neither efficient nor effective for tasks that involve regions.

\begin{figure*}[t]
\centering
\includegraphics[width=1.0\linewidth]{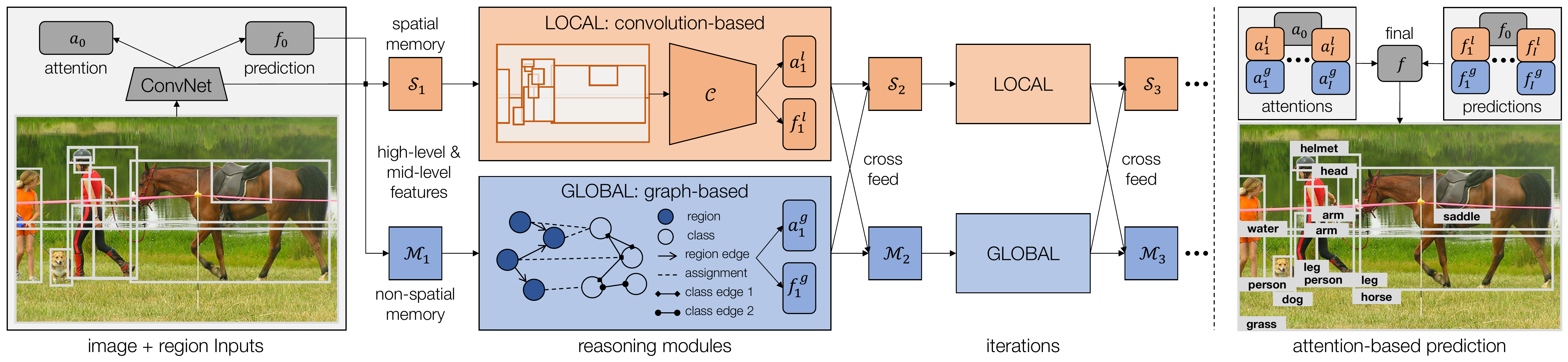}
\caption{{\small Overview of our reasoning framework. Besides a plain ConvNet that gives predictions, the framework has two modules to perform reasoning: a local one (Sec.~\ref{conv}) that uses spatial memory $\mathcal{S}_i$, and reasons with another ConvNet $\mathcal{C}$; and a global one (Sec.~\ref{beyond}) that treats regions and classes as nodes in a graph and reasons by passing information among them. Both modules receive combined high-level and mid-level features, and roll-out iteratively (Sec.~\ref{iter}) while cross-feeding beliefs. The final prediction $f$ is produced by combining all the predictions $f_i$ with attentions $a_i$ (Sec.~\ref{attend}).}\label{fig:overview}}
\vspace{-0.2in}
\end{figure*}

\vspace{-0.05in}
\section{Reasoning Framework}
\vspace{-0.05in}
In this section we build up our reasoning framework. Besides plain predictions $p_0$ from a ConvNet, it consists of two core modules that reason to predict. The first one, local module, uses a spatial memory to store previous beliefs with parallel updates, and still falls within the regime of convolution based reasoning (Sec.~\ref{conv}). Beyond convolutions, we present our key contribution -- a global module that reasons directly between regions and classes represented as nodes in a graph (Sec.~\ref{beyond}). Both modules build up estimation iteratively (Sec.~\ref{iter}), with beliefs cross-fed to each other. Finally taking advantage of both local and global, we combine predictions from all iterations with an attention mechanism (Sec.~\ref{attend}) and train the model with sample re-weighting (Sec.~\ref{train}) that focuses on hard examples (See Fig.~\ref{fig:overview}).

\subsection{Reasoning with Convolutions\label{conv}}
Our first building block, the local module, is inspired from~\cite{chen2017spatial}. At a high level, the idea is to use a spatial memory $\mathcal{S}$ to store previously detected objects at the very location they have been found. $\mathcal{S}$ is a tensor with three dimensions. The first two, height $H$ and width $W$, correspond to the reduced size ($1/16$) of the image. The third one, depth $D$ (${=}512$), makes each cell of the memory $c$ a vector that stores potentially useful information at that location.

$\mathcal{S}$ is updated with both high-level and mid-level features. For high-level, information regarding the estimated class label is stored. However, just knowing the class may not be ideal -- more details about the shape, pose \etc can also be useful for other objects. For example, it would be nice to know the pose of a ``person'' playing tennis to recognize the ``racket''. In this paper, we use the logits $f$ before soft-max activation, in conjunction with feature maps from a bottom convolutional layer $h$ to feed-in the memory. 

Given an image region $r$ to update, we first crop the corresponding features from the bottom layer, and resize it to a predefined square ($7{\times}7$) with bi-linear interpolation as $h$. Since high-level feature $f$ is a vector covering the entire region, we append it to all the $49$ locations. Two $1{\times}1$ convolutions are used to fuse the information~\cite{chen2017spatial} and form our input features $f_r$ for $r$. The same region in the memory $\mathcal{S}$ is also cropped and resized to $7{\times}7$, denoted as $s_r$. After this alignment, we use a convolutional gated recurrent unit (GRU)~\cite{chung2014empirical} to write the memory:
\begin{equation}\label{gru}
    s'_r = u \circ s_r + (1 - u) \circ \sigma(W_f f_r + W_s(z \circ s_r) + b),
\end{equation}
where $s'_r$ is the updated memory for $r$, $u$ is update gate, $z$ is reset gate, $W_f$, $W_s$ and $b$ are convolutional weights and bias, and $\circ$ is entry-wise product. $\sigma(\cdot)$ is an activation function. After the update, $s'_r$ is placed back to $\mathcal{S}$ with another crop and resize operation\footnote{Different from previous work~\cite{chen2017spatial} that introduces an inverse operation to put the region back, we note that crop and resize \emph{itself} with proper extrapolation can simply meet this requirement.}.

\noindent {\bf Parallel Updates.} Previous work~\cite{chen2017spatial} made sequential updates to memory. However, sequential inference is inefficient and GPU-intensive -- limiting it to only give ten outputs per image~\cite{chen2017spatial}. In this paper we propose to update the regions in parallel as an approximation. In overlapping cases, a cell can be covered multiple times from different regions. When placing the regions back to $\mathcal{S}$, we also calculate a weight matrix $\Gamma$ where each entry $\gamma_{r,c}{\in}[0,1]$ keeps track of how much a region $r$ has contributed to a memory cell $c$: $1$ meaning the cell is fully covered by the region, $0$ meaning not covered. The final values of the updated cell is the weighted average of all regions. 

The actual reasoning module, a ConvNet $\mathcal{C}$ of three $3{\times}3$ convolutions and two $4096$-D fully-connected layers, takes $\mathcal{S}$ as the input, and builds connections within the local window of its receptive fields to perform prediction. Since the two-dimensional image structure and the location information is preserved in $\mathcal{S}$, such an architecture is particularly useful for relationships with spatial reasoning.

\begin{figure}[t]
\centering
\includegraphics[width=.8\linewidth]{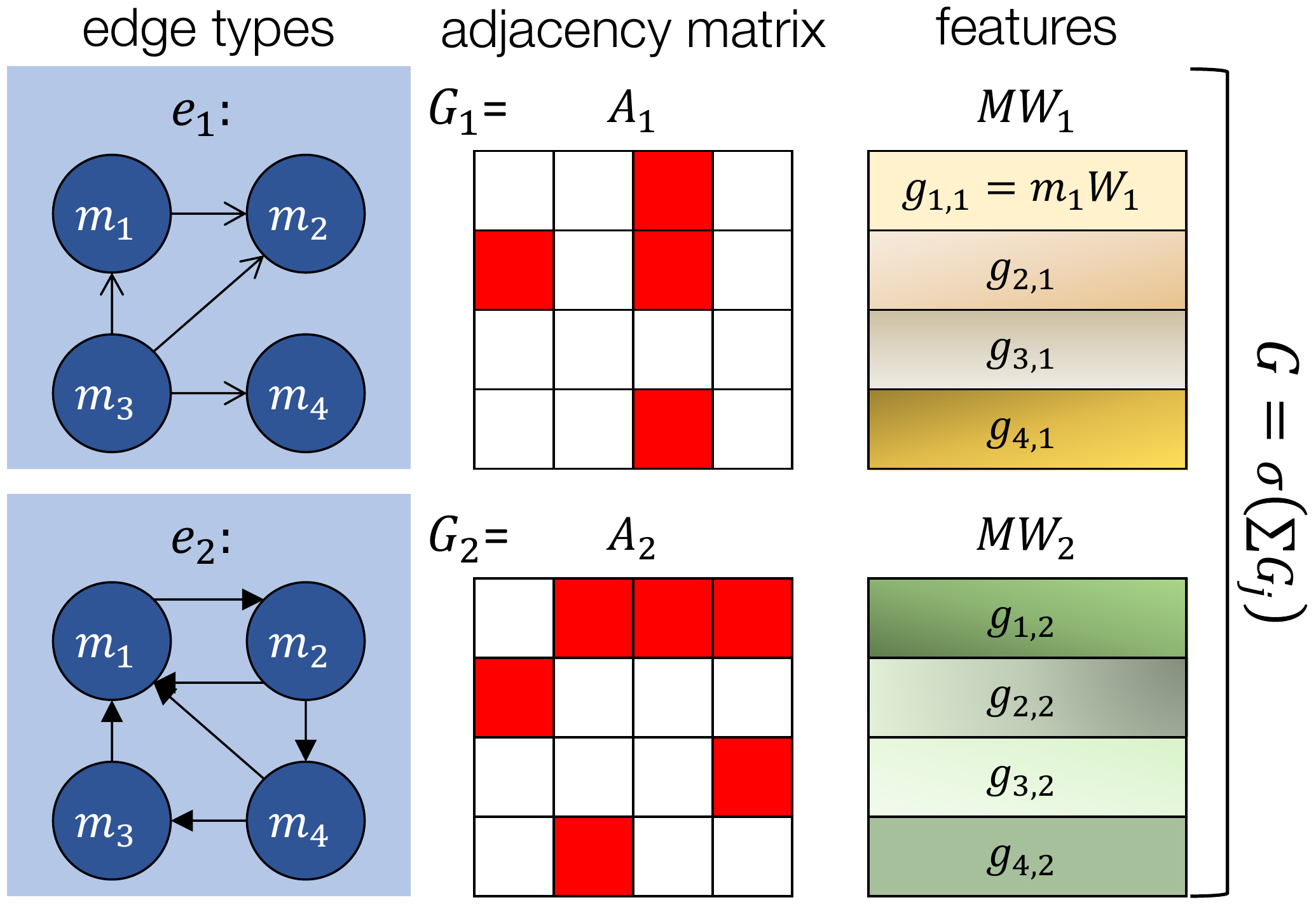}
\caption{{\small Illustration of directly passing information on a graph with multiple edge types. Here four nodes are linked with two edge types. Each node is represented as an input feature vector $m_i$ (aggregated as $M$). Weight matrix $W_j$ is learned for edge type $j$ to transform inputs. Then adjacency matrix $A_j$ is applied to pass information to linked nodes. Finally, output $G$ is generated by accumulating all edge types and apply activation function.}\label{fig:paths}}
\vspace{-0.2in}
\end{figure}

\subsection{Beyond Convolutions\label{beyond}}
Our second module goes beyond local regions and convolutions for global reasoning. Here the meaning of \emph{global} is two-fold. First is \emph{spatial}, that is, we want to let the regions farther away to directly communicate information with each other, not confined by the receptive fields of the reasoning module $\mathcal{C}$. Second is \emph{semantic}, meaning we want to take advantage of visual knowledge bases, which can provide relationships between classes that are globally true (\ie commonsense) across images. To achieve both types of reasoning, we build a graph $\mathcal{G}=(\mathcal{N}, \mathcal{E})$, where $\mathcal{N}$ and $\mathcal{E}$ denote node sets and edge sets, respectively. Two types of nodes are defined in $\mathcal{N}$: region nodes $\mathcal{N}_r$ for $R$ regions, and class nodes $\mathcal{N}_c$ for $C$ classes.  

As for $\mathcal{E}$, three groups of edges are defined between nodes. First for $\mathcal{N}_r$, a spatial graph is used to encode spatial relationships between regions ($\mathcal{E}_{r{\rightarrow}r}$). Multiple types of edges are designed to characterize the relative locations. We begin with basic relationships such as ``left/right'', ``top/bottom'' and we define edge weights by measuring the pixel-level distances between the two. Note that we do not use the raw distance $x$ directly, but instead normalizing it to $[0,1]$ with a kernel $\kappa(x){=}\exp(-x/\Delta)$ (where $\Delta{=}50$ is the bandwidth), with the intuition that closer regions are more correlated. The edge weights are then used directly in the adjacency matrix of the graph. Additionally, we include edges to encode the coverage patterns (\eg intersection over union, IoU~\cite{everingham2010pascal}), which can be especially helpful when two regions overlap. 

A second group of edges lie between regions and classes, where the assignment for a region to a class takes place. Such edges shoulder the responsibility of propagating beliefs from region to class ($e_{r{\rightarrow}c}$) or backwards from class to region ($e_{c{\rightarrow}r}$). Rather than only linking to the most confident class, we choose full soft-max score $p$ to define the edge weights of connections to all classes. The hope that it can deliver more information and thus is more robust to false assignments. 

Semantic relationships from knowledge bases are used to construct the third group of edges between classes ($\mathcal{E}_{c{\rightarrow}c}$). Again, multiple types of edges can be included here. Classical examples are ``is-kind-of'' (\eg between ``cake'' and ``food''), ``is-part-of'' (\eg between ``wheel'' and ``car''), ``similarity'' (\eg between ``leopard'' and ``cheetah''), many of which are universally true and are thus regarded as commonsense knowledge for humans. Such commonsense can be either manually listed~\cite{russakovsky2015imagenet} or automatically collected~\cite{chen2013neil}. Interestingly, even relationships beyond these (\eg actions, prepositions) can help recognition~\cite{marino2016more}. Take ``person ride bike'' as an example, which is apparantly more of an image-specific relationship. However, given less confident predictions of ``person'' and ``bike'', knowing the relationship ``ride'' along with the spatial configurations of the two can also help prune other spurious explanations. To study both cases, we experimented with two knowledge graphs in this paper: one created in-house with mostly commonsense edges, and the other also includes more types of relationships accumulated at a large-scale. For the actual graphs used in our experiments, please see Sec.~\ref{data} for more details.

\begin{figure}[t]
\centering
\includegraphics[width=.9\linewidth]{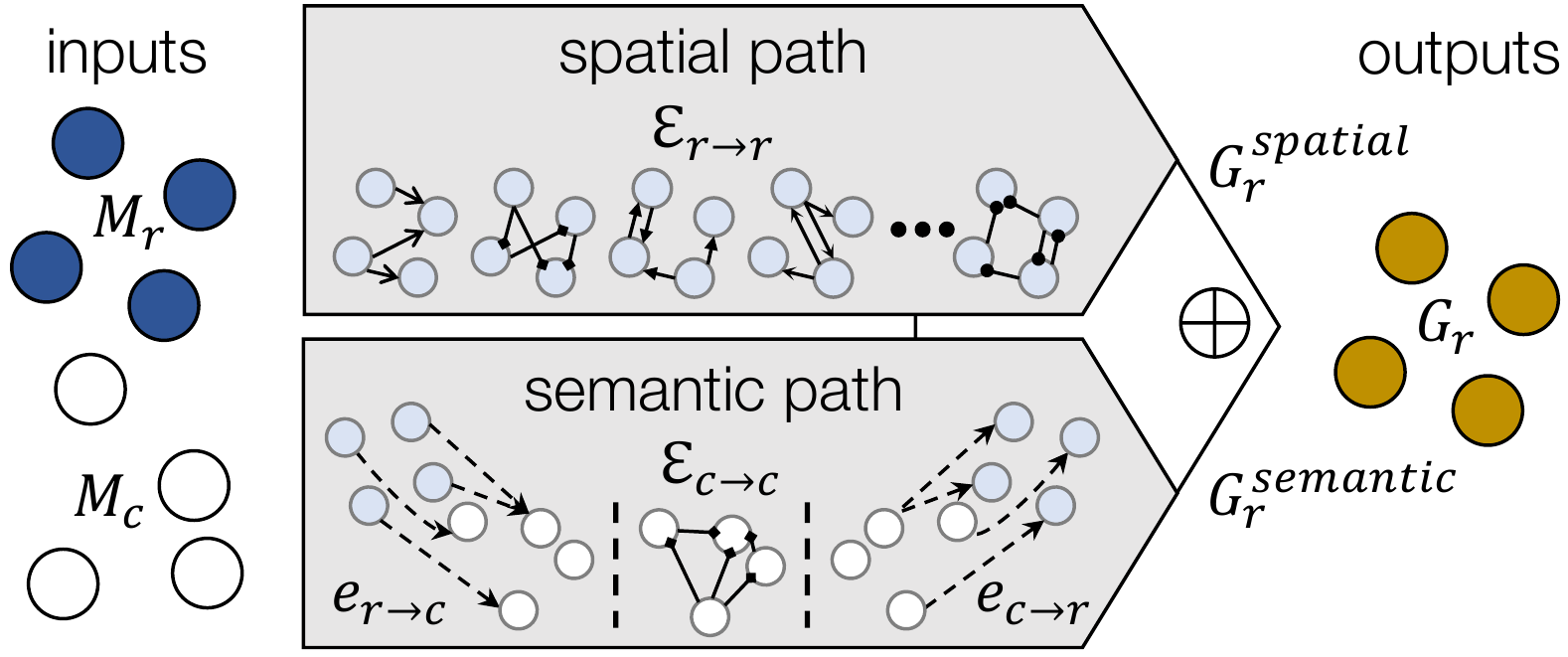}
\caption{{\small Two reasoning paths used in our global reasoning module $\mathcal{R}$. Taking the region and class inputs $M_r$ and $M_c$, the spatial path directly passes information in the region graph with region-to-region edges $\mathcal{E}_{r{\rightarrow}r}$, whereas the semantic path first assigns regions to classes with $e_{r{\rightarrow}c}$, passes the information on to other classes with class-to-class edges $\mathcal{E}_{c{\rightarrow}c}$, and then propagates back. Final outputs are combined to generate output region features $G_r$.}\label{fig:ss}}
\vspace{-0.2in}
\end{figure}

Now we are ready to describe the graph-based reasoning module $\mathcal{R}$. As the input to our graph, we use $M_r{\in}\mathbb{R}^{R\times D}$ to denote the features from all the region nodes $\mathcal{N}_r$ combined, where $D$ (${=}512$) is the number of feature channels. For each class node $n_c$, we choose off-the-shelf word vectors~\cite{joulin2016fasttext} as a convenient representation, denoted as $M_c{\in}\mathbb{R}^{C\times D}$. We then extend previous works~\cite{scarselli2009graph,niepert2016learning} and pass messages directly on $\mathcal{G}$ (See Fig.~\ref{fig:paths}). Note that, because our end-goal is to recognize regions better, all the class nodes should only be used as intermediate ``hops'' for better region representations. With this insight, we design two reasoning paths to learn the output features $G_r$: a \emph{spatial} path on which only region nodes are involved:
\begin{equation}\label{spatial}
    G^{spatial}_r = \sum_{e{\in} \mathcal{E}_{r{\rightarrow}r}}{A_e M_r W_e},
\end{equation}
where $A_e{\in}\mathbb{R}^{r\times r}$ is the adjacency matrix of edge type $e$, $W_e{\in}\mathbb{R}^{d\times d}$ is weight (bias is ignored for simplicity). The second reasoning path is a \emph{semantic} one through class nodes:
\begin{equation}\label{semantic}
    G^{semantic}_c = \sum_{e{\in} \mathcal{E}_{c{\rightarrow}c}}{A_e \sigma(A_{e_{r{\rightarrow}c}} M_r W_{e_{r{\rightarrow}c}} + M_c W_c) W_e},
\end{equation}
where we first map regions to classes through $A_{e_{r{\rightarrow}c}}$ and $W_{e_{r{\rightarrow}c}}$, combine the intermediate features with class features $M_c$, and again aggregate features from multiple types of edges between classes.
Finally, the output for regions $G_r$ are computed by merging these two paths:
\begin{equation}\label{output}
    G_r = \sigma(G^{spatial}_r + \sigma(A_{e_{c{\rightarrow}r}} G^{semantic}_c W_{e_{c{\rightarrow}r}})),
\end{equation}
which first propagates semantic information back to regions, and then applies non-linear activation (See Fig.~\ref{fig:ss}).

Just like convolution filters, the above-described paths can also be stacked, where the output $G_r$ can go through another set of graph operations -- allowing the framework to perform joint spatial-semantic reasoning with deeper features. We use three stacks of operations with residual connections~\cite{he2016deep} in $\mathcal{R}$, before the output is fed to predict.

\subsection{Iterative Reasoning\label{iter}}
A key ingredient of reasoning is to iteratively build up estimates. But how does information pass from one iteration to another? Our answer is \emph{explicit} memory, which stores all the history from previous iterations. The local module uses spatial memory $\mathcal{S}$, and the global module uses another memory $\mathcal{M}$ but without spatial structures. At iteration $i$, $\mathcal{S}_i$ is followed by convolutional reasoning module $\mathcal{C}$ to generate new predictions $f_i^l$ for each region. Similarly, global module also gives new predictions $f_i^g$ from $\mathcal{R}$. These new predictions as high-level features can then be used to get the updated memories $\mathcal{S}_{i+1}$ and $\mathcal{M}_{i+1}$. The new memories will lead to another round of updated $f_{i+1}$s and the iteration goes on. 

While one can do local and global reasoning in isolation, both the modules work best in conjunction. Therefore, for our full pipeline we want to join force of both modules when generating the predictions. To this end, we introduce \emph{cross-feed} connections. After reasoning, both the local and global features are then concatenated together to update the memories $\mathcal{S}_{i+1}$ and $\mathcal{M}_{i+1}$ using GRU. In this way, spatial memory can benefit from global knowledge of spatial and semantic relationships, and graph can get a better sense of the local region layouts. 

\subsection{Attention\label{attend}}
Inspired from the recent work on attention~\cite{chen2016attention}, we make another modification at the model output. Specifically, instead of only generating scores $f$, the model also has to produce an ``attention'' value $a$ that denotes the relative confidence of the current prediction compared to the ones from other iterations or modules. Then the fused output is a weighted version of all predictions using attentions. Mathematically, if the model roll-outs $I$ times, and outputs $N{=}2I{+}1$ (including $I$ local, $I$ global and $1$ from plain ConvNet) predictions $f_n$, using attentions $a_n$, the final output $f$ is calculated as:
\begin{equation}\label{att}
    f = \sum_{n}{w_n f_n}, \quad\mathrm{where}\quad w_n=\frac{\exp(-a_n)}{\sum_{n'}{\exp(-a_{n'})}}.
\end{equation}
Note again that here $f_n$ is the logits before soft-max, which is then activated to produce $p_n$. The introduction of attention allows the model to intelligently choose feasible predictions from different modules and iterations.

\subsection{Training\label{train}}
Finally, the overall framework is trained end-to-end, with a total loss function consists of: a) plain ConvNet loss $\mathcal{L}_{0}$; b) local module loss $\mathcal{L}^l_{i}$; c) global module loss $\mathcal{L}^g_{i}$; and d) the final prediction loss with attentions $\mathcal{L}_f$.

Since we want our reasoning modules to focus more on the harder examples, we propose to simply \emph{re-weight} the examples in the loss, based on predictions from previous iterations. Formally, for region $r$ at iteration $i{\ge}1$, the cross-entropy loss for both modules is computed as: 
\begin{equation}\label{reweight}
    \mathcal{L}_{i}(r) = \frac{\max(1. - p_{i-1}(r), \beta)}{\sum_{r'}\max(1. - p_{i-1}(r'), \beta)}\log(p_{i}(r)),
\end{equation}
where $p_{i}(r)$ is the soft-max output of the ground-truth class, and $\beta{\in}[0,1]$ controls the entropy of the weight distribution: when $\beta{=}1$, it is uniform distribution; and when $\beta{=}0$, entropy is minimized. In our experiments, $\beta$ is set to $0.5$. $p_{i-1}(r)$ is used as features without back-propagation. For both local and global, $p_{0}(r)$ is the output from the plain ConvNet. 

\vspace{-0.05in}
\section{Experiments}
\vspace{-0.05in}
In this section we evaluate the effectiveness of our framework. We begin with our experimental setups, which includes the datasets to work with (Sec.~\ref{data}), the task to evaluate on (Sec.~\ref{task}) and details of our implementation (Sec.~\ref{details}). We discuss our results and analyze them in Sec.~\ref{results} and Sec.~\ref{ablative} respectively. 

\subsection{Datasets and Graphs\label{data}}
Datasets are biased~\cite{torralba2011unbiased}. For context reasoning we would naturally like to have scene-focused datasets~\cite{zhou2016semantic} as opposed to object-focused ones~\cite{russakovsky2015imagenet}. To showcase the capabilities of our system, we need densely labeled dataset with a large number of classes. Finally, one benefit of using knowledge graph is to transfer across classes, therefore a dataset with \emph{long-tail} distribution is an ideal test-bed. Satisfying all these constraints, ADE~\cite{zhou2016semantic} and Visual Genome (VG)~\cite{krishna2016visual} where regions are densely labeled in open vocabulary are the main picks of our study. 

For ADE, we use the publicly released training set ($20,210$) images for training, and split the validation set ($2,000$ images) into {\tt val-1k} and {\tt test-1k} with $1,000$ images each. The original raw names are used due to a more detailed categorization~\cite{zhou2016semantic}. We filter out classes with less than five instances, which leaves us with $1,484$ classes. With the help of parts annotations in the dataset, a commonsense knowledge graph is created with five types of edges between classes: a) ``is-part-of'' (\eg ``leg'' and ``chair''); b) ``is-kind-of'' (\eg ``jacket'' and ``clothes''); c) ``plural-form'' (\eg ``tree'' and ``trees''); d) ``horizontal-symmetry'' (\eg ``left-arm'' and ``right-arm''); e) ``similarity'' (\eg ``handle'' and ``knob''). Notice that the first four types are directed edges, hence we also include their inverted versions. 

For VG, the latest release (v$1.4$) is used. We split the entire set of $108,077$ images into $100$K, $4,077$ and $4$K as {\tt train}, {\tt val} and {\tt test} set. Similar pre-processing is done on VG, except that we use synsets~\cite{russakovsky2015imagenet} instead of raw names due to less consistent labels from multiple annotators. $3,993$ classes are used. For knowledge graph between classes, we take advantage of the relationship annotations in the set, and select the top $10$ most frequent relationships to automatically construct edges beyond commonsense relationships constructed for ADE. For each type of relationships, the edge weights are normalized so that each row of the adjacency matrix is summed-up to one. While this approach results in a noisier graph, it also allows us to demonstrate that our approach is scalable and robust to noise.

Finally, we also show experiments on COCO~\cite{lin2014microsoft}. However, since it is detection oriented -- has only $80$ classes picked to be mutually-exclusive, and covers less percentage of labeled pixels, we only report results a) without the knowledge graph and b) without a test split ({\tt trainval35k}~\cite{chen2017spatial} for training and {\tt minival} for evaluation). This setup is for analysis purposes only.

\begin{table}[t]
\centering
\renewcommand{\arraystretch}{1.1}
\renewcommand{\tabcolsep}{1.2mm}
\caption{\label{tab:final}{Main results on ADE {\tt test-1k} and VG {\tt test}. AP is average precision, AC is classification accuracy. Superscripts show the improvement $\nabla$ over the baseline.}}
\resizebox{1.0\linewidth}{!}{
\begin{tabular}{@{} C{0.5cm} !{\vrule} L{2.5cm} !{\vrule} x{1.2cm} x{1.2cm} !{\vrule} x{1.2cm} x{1.2cm} @{}}
\Xhline{1pt}
\multirow{2}{*}{$\%$} & \multirow{2}{*}{\textbf{Method}} & \multicolumn{2}{c!{\vrule}}{per-instance} & \multicolumn{2}{c}{per-class} \\
\Xcline{3-6}{0.5pt}
& & AP\textsuperscript{$\nabla$} & AC\textsuperscript{$\nabla$} & AP\textsuperscript{$\nabla$} & AC\textsuperscript{$\nabla$} \\
\Xhline{1pt}
\parbox[t]{2.5mm}{\multirow{7}{*}{\rotatebox[origin=c]{90}{\small ADE}}} & Baseline & 67.0 & 67.0 & 40.1 & 33.2 \\
& ~~~~{\small w/ ResNet-101} & 68.2 & 68.3 & 40.8 & 34.4 \\
& ~~~~{\small w/ $800$-input} & 68.2 & 68.2 & 41.0 & 34.3 \\
& ~~~~{\small Ensemble} & 68.7 & 68.8 & 42.9 & 35.3 \\
\Xcline{2-6}{0.5pt}
& Ours\textsubscript{-Local} & 71.6\textsuperscript{+4.6} & 71.7\textsuperscript{+4.7} & 47.9\textsuperscript{+7.8} & 38.7\textsuperscript{+5.7} \\
& Ours\textsubscript{-Global} & 69.8\textsuperscript{+2.8} & 69.8\textsuperscript{+2.8} & 44.5\textsuperscript{+4.4} & 36.8\textsuperscript{+3.6} \\
& Ours\textsubscript{-Final} & {\bf 72.6}\textsuperscript{+5.6} & {\bf 72.6}\textsuperscript{+5.6} & {\bf 48.5}\textsuperscript{+8.4} & {\bf 39.5}\textsuperscript{+6.3} \\

\Xhline{0.5pt}
\parbox[t]{2.5mm}{\multirow{7}{*}{\rotatebox[origin=c]{90}{\small VG}}} & Baseline & 49.1 & 49.6 & 16.9 & 12.1 \\
& ~~~~{\small w/ ResNet-101} & 50.3 & 50.8 & 18.0 & {\bf 13.0} \\
& ~~~~{\small w/ $800$-input} & 49.5 & 50.0 & 17.0 & 12.2 \\
& ~~~~{\small w/ Ensemble} & 50.2 & 50.7 & 17.7 & 12.3 \\ 
\Xcline{2-6}{0.5pt}
& Ours\textsubscript{-Local} & 51.4\textsuperscript{+2.3} & 51.9\textsuperscript{+2.3} & 18.8\textsuperscript{+1.9} & 12.8\textsuperscript{+0.7} \\
& Ours\textsubscript{-Global} & 50.9\textsuperscript{+1.8} & 51.5\textsuperscript{+1.9} & 18.3\textsuperscript{+1.4} & 12.6\textsuperscript{+0.5} \\
& Ours\textsubscript{-Final} & {\bf 51.7}\textsuperscript{+2.6} & {\bf 52.2}\textsuperscript{+2.6} & {\bf 19.1}\textsuperscript{+2.2} & 12.9\textsuperscript{+0.8} \\

\Xhline{1pt}
\end{tabular}
}
\vspace{-0.2in}
\end{table}

\begin{figure*}[t]
\centering
\includegraphics[width=1.0\linewidth]{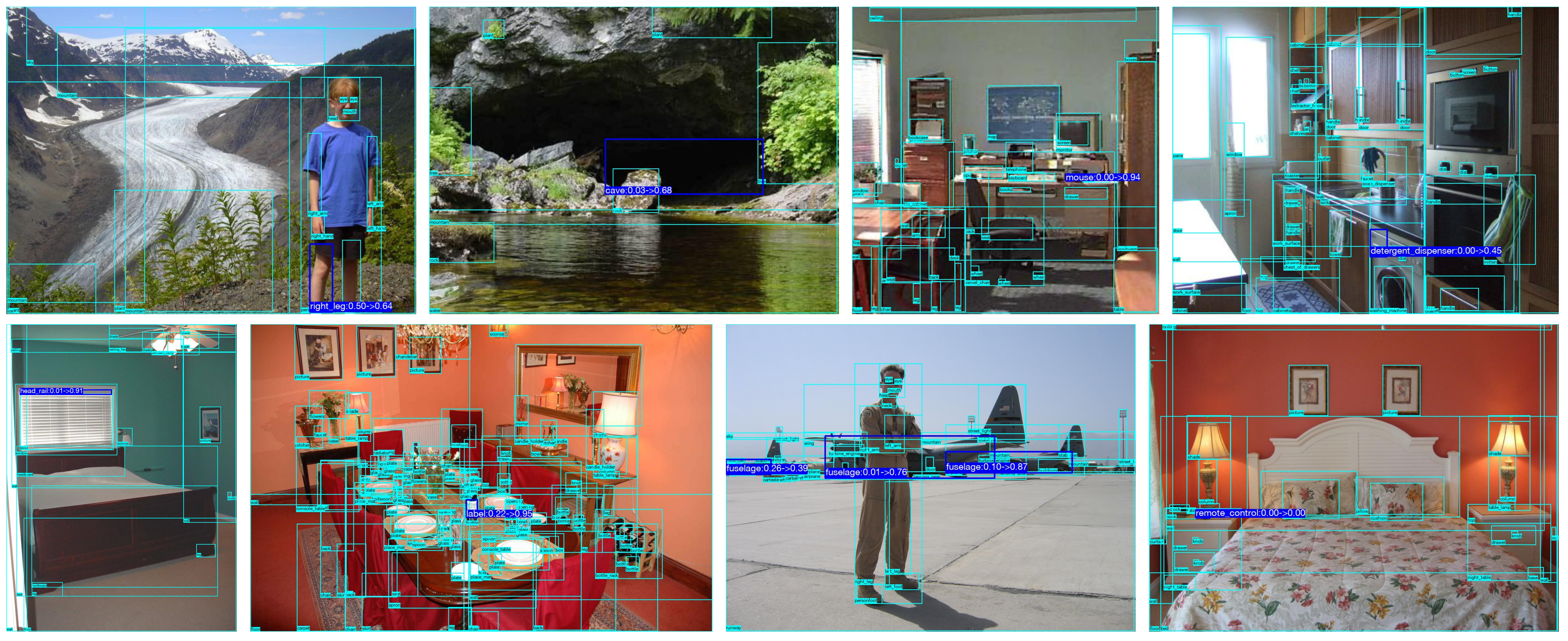}
\caption{{\small Qualitative examples from ADE {\tt test-1k} (best if zoomed-in). For regions highlighted in blue, the predictions from baseline and our model are compared. Other regions are also listed to provide the context. For example, the ``right-leg'' is less confused with ``left-leg'' after reasoning (top-left); the ``mouse'' on the ``desk'' is predicted despite low resolution (top-third); and ``detergent-dispenser'' is recognized given the context of ``washing-machine'' (top-right). At bottom-right we show a failure case where context does not help ``remote-control'', probably because it has never appeared on the ``night-table'' before and no semantic relationship is there to help.}\label{fig:examples}}
\vspace{-0.2in}
\end{figure*}

\subsection{Task and Evaluation\label{task}}
We evaluate our system on the task of region classification, where the goal is to assign labels to designated regions denoted by rectangular bounding boxes. For both training and testing, we use provided ground-truth locations. We picked this task for three reasons. The {\bf first} one is on evaluation. As the number of classes increases in the vocabulary, \emph{missing} labels are inevitable, which is especially severe for object parts (\eg ``rim'', ``arm'') and related classes (\eg ``shoes'' \vs ``sneakers'') where external knowledge is valuable. If there are missing labels, fair evaluation becomes much more difficult since accuracy becomes impossible to evaluate -- cannot tell if a prediction is wrong, or the label itself is missing. Interestingly, such an issue also happens to other research areas (\eg recommendation systems~\cite{sarwar2001item} and link prediction~\cite{liben2007link}). Borrowing ideas from them, a practical solution is to evaluate \emph{only} on what we already know -- in our case ground-truth regions. {\bf Second}, although region classification is a simplified version of object detection and semantic segmentation, it maintains a richer set of labels, especially including ``stuff'' classes like ``road'', ``sky'', and object instances. Modeling ``stuff-object'' and instance-level relationships is a crucial capability which would be missed in a pure detection/segmentation setting. {\bf Finally} as our experiment will show (Sec.~\ref{ablative}), while object detectors can be used off-the-shelf, the additional manually defined parameters and components (\eg overlapping threshold for a region to be positive/negative, predefined scale/aspect ratio sets of anchors~\cite{ren2015faster}) in its pipeline pose limitations on how much context can benefit. For example, after non-maximal suppression (NMS), highly overlapping objects (\eg ``window'' and ``shutter'') will be suppressed, and ironically this is exactly where context reasoning could have helped. On the other hand, by feeding fixed regions directly for end-to-end learning, we can at least factorize the \emph{recognition} error from the \emph{localization} one~\cite{hoiem2012diagnosing}, and get a clean focus on how context can help discriminating confusing classes.

Since ADE is a segmentation dataset, we convert segmentation masks to bounding boxes. For object classes (\eg ``person''), each instance is created a separate box. Part (\eg ``head'') and part-of-part (\eg ``nose'') are also included. For VG and COCO, boxes are directly used.

For evaluation, we use classification accuracy (AC) and average precision (AP)~\cite{everingham2010pascal}. Note that since all the regions are fixed with known labels, there is no need to set a region overlap threshold for AP. Results can be aggregated in two ways: the first way (``per-class'') computes metrics separately for each class in the set, and take the mean; since the final scores are all taken from a calibrated soft-max output, a second way (``per-instance'') that computes metrics simultaneously for all classes. Intuitively, ``per-class'' assigns more weights to instances from rare classes.

\subsection{Implementation Details\label{details}}
A simplified version of \emph{tf-faster-rcnn}\footnote{\url{https://github.com/endernewton/tf-faster-rcnn}} is used to implement our baseline for region classification, with region proposal branch and bounding box regression components removed. Unless otherwise noted, ResNet-50~\cite{he2016deep} pre-trained on ImageNet~\cite{russakovsky2015imagenet} is used as our backbone image classifier, and images are enlarged to shorter size $600$ pixels during both training and testing. Specifically, full-image shared convolutional feature maps are computed till the last \emph{conv4} layer. Then the ground-truth boxes are used as regions-of-interest to compute region-specific features (crop and resize to $7{\times}7$ without max-pool). All layers of \emph{conv5} and up are then adopted to obtain the final feature for the baseline prediction $p_0$. Batch normalization parameters are fixed.

For the local module, we use the last \emph{conv4} layer as our mid-level features to feed the spatial memory $\mathcal{S}$. For the global module, mid-level features are the final \emph{conv5} ($2048$-D) layer after avg-pool. Both features are fused with the logits before soft-max $f$, and then fed into the memory cells. Word vectors from fastText~\cite{joulin2016fasttext} are used to represent each class, which extracts sub-word information and generalizes well to out-of-vocabulary words. ReLU is selected as the activation function. We roll-out the reasoning modules $3$ times and concurrently update all regions at each iteration, as more iterations do not offer more help.

We apply stochastic gradient descent with momentum to optimize all the models, and use the validation set to tune hyper-parameters. Our final setups are: $5e^{-4}$ as the initial learning rate, reduced once ($0.1{\times}$) during fine-tuning; $1e^{-4}$ as weight decay; $0.9$ as momentum. For ADE, we train $320$K iterations and reduce learning rate at $280$K. For VG and COCO the numbers are $640$K/$500$K and $560$K/$320$K, respectively\footnote{Training longer still reduces cross-entropy, but drops both AP and AC.}. We use a single image per step, and the only data augmentation technique used during training is left-right flipping\footnote{The labels for class pairs like ``left-hand'' and ``right-hand'' are swapped for flipped images.}. No augmentation is used in testing. 

\subsection{Main Results\label{results}}
Quantitative results on ADE {\tt test-1k} and VG {\tt test} are shown in Tab.~\ref{tab:final}. Besides plain ConvNet $p_0$, we also add three more baselines. First, we use ResNet-101 as the backbone to see the performance can benefit from deeper networks. Second, we increase the input image size with a shorter side $800$ pixels, which is shown helpful especially for small objects in context~\cite{lin2016feature}. Finally, to check whether our performance gain is a result of more parameters, we include model ensemble as the third baseline where the prediction of two separate baseline models are averaged.

As can be seen, our reasoning modules are performing much better than all the baselines on ADE. The local module alone can increase per-class AP by $7.8$ absolute points. Although the global module alone is not as effective ($4.4\%$ improvement), the performance gain it offers is \emph{complementary} to the local module, and combining both modules we arrive at an AP of $48.5\%$ compared to the baseline AP $40.1\%$. On the other hand, deeper network and larger input size can only help ${\sim}1\%$, less than model ensembles. Additionally, our models achieve higher per-class metric gains than per-instance ones, indicating that \emph{rare} classes get helped more -- a nice property for learning from few examples. Some qualitative results are listed in Fig.~\ref{fig:examples}. 

We also report the speed for future reference. On Titan Xp, the final model on ADE trains at 0.344s per iteration, compared to the baseline ResNet-50 at $0.163$s and ResNet-101 at $0.209$s. For testing, our model takes $0.165$s, whereas ResNet-50 $0.136$s, ResNet-101 $0.156$s. We believe the additional
cost is minimal with regard to the extra accuracy.

We see a similar but less significant trend on VG. This can potentially be a result of \emph{noisier} labels -- for ADE (and COCO shown later), the per-instance AP and AC values are within $0.1\%$, intuitively suggesting that \emph{higher} scores usually correspond to correct classifications. However, on VG the difference is at ${\sim}0.5\%$, meaning more of the highly confident predictions are not classified right, which are likely caused by missing ground-truths. 

\subsection{Analysis\label{ablative}}
Our analysis is divided into two major parts. In the first part, we conduct thorough ablative analysis on the framework we have built. Due to space limitation, we only report results on ADE here at Tab.~\ref{tab:contribute}, for more analysis on VG, please check our supplementary material. 

As can be seen, re-weighting hard examples with Eq.~\ref{reweight} helps around $0.5\%$ regardless of reasoning modules. Spatial memory $\mathcal{S}$ is critical in the local module -- if replaced by feeding last \emph{conv4} layer directly the performance drops almost to baseline. Local context aggregator $\mathcal{C}$ is less influential for ADE since the regions including background are densely labeled. A different story takes place at the global module: removing the reasoning module $\mathcal{R}$ steeply drops performance, whereas further removing memory $\mathcal{M}$ does not hurt much. Finally, for our full pipeline, removing cross-feeding and dropping the number of iterations both result in worse performance.

\begin{table}[t]
\centering
\renewcommand{\arraystretch}{1.1}
\renewcommand{\tabcolsep}{1.2mm}
\definecolor{LightGreen}{rgb}{0.75,1,0.75}
\definecolor{LightRed}{rgb}{1,0.75,0.75}
\definecolor{LightBlue}{rgb}{0.75,0.75,1}
\caption{\label{tab:contribute}{Ablative analysis on ADE {\tt test-1k}. In the first row of each block we repeat Local, Global and Final results from Tab.~\ref{tab:final}. Others see Sec.~\ref{ablative} for details.}}
\resizebox{1.0\linewidth}{!}{
\begin{tabular}{@{} C{0.5cm} !{\vrule} L{2.5cm} !{\vrule} x{1.2cm} x{1.2cm} !{\vrule} x{1.2cm} x{1.2cm} @{}}
\Xhline{1pt}
\multirow{2}{*}{$\%$} & \multirow{2}{*}{\textbf{Analysis}} & \multicolumn{2}{c!{\vrule}}{per-instance} & \multicolumn{2}{c}{per-class} \\
\Xcline{3-6}{0.5pt}
& & AP & AC & AP & AC \\
\Xhline{1pt}
\parbox[t]{2.5mm}{\multirow{4}{*}{\rotatebox[origin=c]{90}{\small Local}}} & Ours\textsubscript{-Local} & 71.6 & 71.7 & 47.9 & 38.7 \\
& ~~~~{\small w/o re-weight} & 71.3 & 71.3 & 46.7 & 37.9 \\
& ~~~~{\small w/o $\mathcal{C}$} & 70.9 & 71.0 & 46.1 & 37.5 \\
& ~~~~{\small w/o $\mathcal{S}$} & 67.6 & 67.6 & 42.1 & 34.4 \\

\Xhline{0.5pt}
\parbox[t]{2.5mm}{\multirow{6}{*}{\rotatebox[origin=c]{90}{\small Global}}} & Ours\textsubscript{-Global} & 69.8 & 69.8 & 44.5 & 36.8 \\
& ~~~~{\small w/o re-weight} & 69.2 & 69.2 & 43.8 & 36.7 \\
& ~~~~{\small w/o spatial} & 67.8 & 67.8 & 41.5 & 35.0 \\
& ~~~~{\small w/o semantic} & 69.1 & 69.2 & 43.9 & 35.9 \\
& ~~~~{\small w/o $\mathcal{R}$} & 67.1 & 67.2 & 41.5 & 34.5 \\
& ~~~~{\small w/o $\mathcal{M}$ \& $\mathcal{R}$} & 67.1 & 67.1 & 41.0 & 34.0 \\

\Xhline{0.5pt}
\parbox[t]{2.5mm}{\multirow{4}{*}{\rotatebox[origin=c]{90}{\small Final}}} & Ours\textsubscript{-Final} & 72.6 & 72.6 & 48.5 & 39.5 \\
& ~~~~{\small w/o re-weight} & 72.1 & 72.2 & 47.3 & 38.6 \\
& ~~~~{\small w/o cross-feed} & 72.2 & 72.2 & 47.6 & 39.0 \\
& ~~~~{\small $2$ iterations} & 71.9 & 72.0 & 48.1 & 39.0 \\

\Xhline{1pt}
\end{tabular}
}
\vspace{-0.1in}
\end{table}

\noindent{\bf Missing Regions.} So far we have shown results when all the regions are present. Next, we want to analyze if our framework is robust to missing regions: if some percentage of regions are not used for reasoning. This will be a common scenario if we use our framework in the detection setting -- the underlying region proposal network~\cite{ren2015faster} may itself miss some regions. We perform this set of experiments on COCO, since its regions are object-focused.

We test three variations. In the first variation, the same region classification pipeline is applied as-is. In the other two, we drop regions. While we could have done it randomly, we simulate the real-world scenario by using region proposals from faster R-CNN~\cite{ren2015faster} ($1190$K/$900$K, {\tt minival} detection mAP $32.4\%$) for testing, where $300$ region proposals after NMS are applied to filter the ground-truth regions (max IoU${>}\delta$). Evaluation is only done on the remaining regions. Here we choose not to use region proposals directly, since the model has seen ground truth regions only. We test two variations: a) ``pre'', where the regions are filtered before inference, \ie only the remaining ground-truths are fed for reasoning; ``post'', where regions are filtered after inference. Note that for the baseline, ``pre'' and ``post'' makes no difference performance-wise.

\begin{table}[t]
\centering
\renewcommand{\arraystretch}{1.1}
\renewcommand{\tabcolsep}{1.2mm}
\caption{\label{tab:coco}{Results with missing regions when region proposals are used. COCO {\tt minival} is used since it is more detection oriented. {\bf pre} filters regions before inference, and {\bf post} filters after inference.}}
\resizebox{1.0\linewidth}{!}{
\begin{tabular}{@{} L{1.8cm} !{\vrule} C{0.6cm} C{0.6cm}  !{\vrule} x{1.2cm} x{1.2cm} !{\vrule} x{1.2cm} x{1.2cm} @{}}
\Xhline{1pt}
\multirow{2}{*}{\textbf{Method}} & \multirow{2}{*}{\bf \small pre} & \multirow{2}{*}{\bf \small post} & \multicolumn{2}{c!{\vrule}}{per-instance} & \multicolumn{2}{c}{per-class} \\
\Xcline{4-7}{0.5pt}
& & & AP\textsuperscript{$\nabla$} & AC\textsuperscript{$\nabla$} & AP\textsuperscript{$\nabla$} & AC\textsuperscript{$\nabla$} \\
\Xhline{1pt}
Baseline & & & 83.2 & 83.2 & 83.7 & 75.9 \\
Ours\textsubscript{-Local} & & & 84.9\textsuperscript{+1.7} & 84.9\textsuperscript{+1.7} & 85.8\textsuperscript{+2.1} & 77.6\textsuperscript{+1.7} \\
Ours\textsubscript{-Global} & & & 85.6\textsuperscript{+2.4} & 85.7\textsuperscript{+2.5} & 86.9\textsuperscript{+3.2} & 78.2\textsuperscript{+2.3} \\
Ours\textsubscript{-Final} & & & {\bf 86.0}\textsuperscript{+2.8} & {\bf 86.0}\textsuperscript{+2.8} & {\bf 87.4}\textsuperscript{+3.7} & {\bf 79.0}\textsuperscript{+3.1} \\
\Xhline{0.5pt}
Baseline & - & - & 87.0 & 87.0 & 87.7 & 80.2 \\
Ours\textsubscript{-Final} & \cmark & & 88.6\textsuperscript{+1.6} & 88.6\textsuperscript{+1.6} & 89.9\textsuperscript{+2.2} & {\bf 82.6}\textsuperscript{+2.4} \\
Ours\textsubscript{-Final} & & \cmark & {\bf 88.8}\textsuperscript{+1.8} & {\bf 88.8}\textsuperscript{+1.8} & {\bf 90.1}\textsuperscript{+2.4} & 82.5\textsuperscript{+2.3} \\
\Xhline{1pt}
\end{tabular}
}
\vspace{-0.1in}
\end{table}

\begin{figure}[t]
\centering
\includegraphics[width=1.\linewidth]{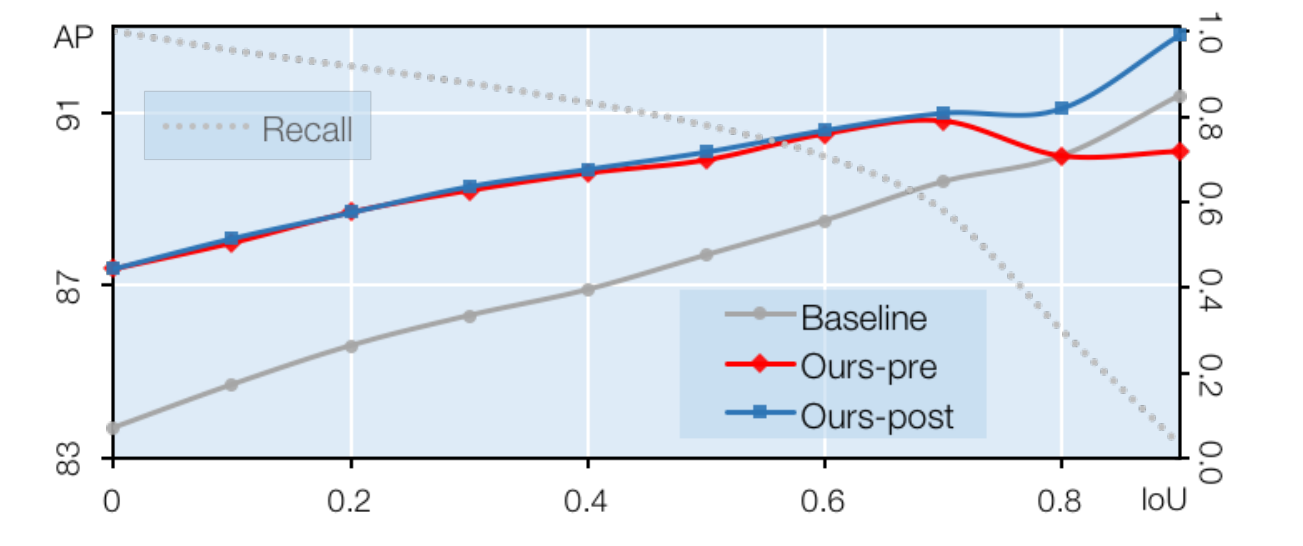}
\caption{{\small Trends of recall and per-class AP when varying IoU threshold $\delta$ from $0$ to $.9$ to drop regions. See text for details.}\label{fig:trend}}
\vspace{-0.2in}
\end{figure}

The results are summarized in Tab.~\ref{tab:coco}. Interestingly, despite lacking a knowledge graph, our global module works better than the local module with the region graph alone, likely due to its power that allows direct region-to-region communication even for farther-away pairs. Combining the two, we report $3.7\%$ absolute advantage on per-class AP over the baseline even with all classes being objects -- no ``stuff'' classes involved.

In Fig.~\ref{fig:trend}, we vary $\delta$ from $0$ to $.9$: with $0$ keeping all regions and $0.9$ dropping the most. As the trend shows, while the reasoning module suffers when regions are dropped, it is quiet resilient and the performance degradation is smooth. For example (listed in Tab.~\ref{tab:coco}), with an IoU threshold $\delta$ of $0.5$ that recalls $78.1\%$ of the ground truth boxes, we still outperform the baseline by $2.4\%$ in the ``post'' setting, and $2.2\%$ in ``pre'' where not all regions can be fed for reasoning. The lower gap implies a) region proposals are usually corresponding to easy examples where less context is needed, and b) context reasoning frameworks like ours benefit from more known regions. At $\delta{=}.8$ the recall ($30.5\%$) is so small that it cannot afford much reasoning, and at $\delta{=}.9$ (recall $3.9\%$), reasoning even hurts the performance.

\vspace{-0.05in}
\section{Conclusion}
\vspace{-0.05in}
We presented a novel framework for iterative visual reasoning. Beyond convolutions, it uses a graph to encode spatial and semantic relationships between regions and classes and passes message on the graph. We show strong performance over plain ConvNets, \eg achieving an $8.4\%$ absolute gain on ADE and $3.7\%$ on COCO. Analysis also shows that our reasoning framework is resilient to missing regions caused by current region proposal approaches. 

\noindent {\bf Acknowledgements}: This work was supported in part by ONR MURI N000141612007. XC would also like to thank Shengyang Dai and Google Cloud AI team for support during the internship. 

{\small
\bibliographystyle{ieee}

}

\end{document}